\documentclass{article}

\usepackage{arxiv}

\usepackage[utf8]{inputenc} 
\usepackage[T1]{fontenc}    
\usepackage{hyperref}       
\usepackage{url}            
\usepackage{booktabs}       
\usepackage{amsfonts}       
\usepackage{nicefrac}       
\usepackage{microtype}      
\usepackage{lipsum}         
\usepackage{graphicx}
\usepackage[numbers,sort&compress]{natbib}
\usepackage{doi}
\usepackage{mathtools}
\usepackage{cleveref}       
\usepackage{xcolor}

\title{Using t-distributed stochastic neighbor embedding for visualization and segmentation of 3D point clouds of plants}

\author{ \href{https://orcid.org/0000-0000-0000-0000}{\includegraphics[scale=0.06]{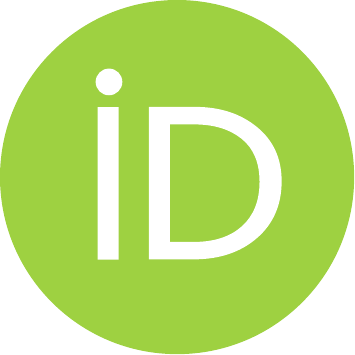}\hspace{1mm}Helin Dutagaci}\thanks{hdutagaci@ogu.edu.tr, helindutagaci@gmail.com} \\
	Department of Electrical-Electronics Engineering\\
	Osmangazi University\\
	Eskişehir \\
	\texttt{hdutagaci@ogu.edu.tr} \\
}

\renewcommand{\shorttitle}{t-SNE for 3D point clouds of plants}

\hypersetup{
pdftitle={Using t-distributed stochastic neighbor embedding for visualization and segmentation of 3D point clouds of plants},
pdfsubject={t-SNE for 3D segmentation},
pdfauthor={Helin Dutagaci},
pdfkeywords={t-SNE, Point Clouds, Plants, Visualization, Superpoint, Semantic Segmentation, Instance Segmentation, Phenotyping},
}

\begin{document}
\maketitle

\begin{abstract}
	In this work, the use of t-SNE is proposed to embed 3D point clouds of plants into 2D space for plant characterization. It is demonstrated that t-SNE operates as a practical tool to flatten and visualize a complete 3D plant model in 2D space. The perplexity parameter of t-SNE allows 2D rendering of plant structures at various organizational levels. Aside from the promise of serving as a visualization tool for plant scientists, t-SNE also provides a gateway for processing 3D point clouds of plants using their embedded counterparts in 2D. In this paper, simple methods were proposed to perform semantic segmentation and instance segmentation via grouping the embedded 2D points. The evaluation of these methods on a public 3D plant data set conveys the potential of t-SNE for enabling of 2D implementation of various steps involved in automatic 3D phenotyping pipelines.
\end{abstract}


\section{Introduction}

Automatic trait estimation of plants is becoming an indispensable component for many applications such as crop monitoring \citep{yang2017unmanned}, crop quality assessment \citep{Garbez2018}, agricultural robotics \citep{roldan2018robots}, and phenotyping \citep{Minervini2015, James2022}. The developments in 3D reconstruction technologies and 3D computer vision tools allow 3D modeling of plants and their automatic trait estimation on digitized 3D models. Examples to such traits are leaf area, leaf length, leaf inclination angle, stem height, internode lengths and branching angles \citep{paulus2019measuring}. 

Characterization of 3D plants poses a particular challenge due to high structural heterogeneity, high self-occlusion, and high structural complexity. The complexity in the connectivity of parts, various parts occluding each other, and variation among parts cause difficulty for both manual and automatic inspection and deliniation of the organs. Despite this challenge, 3D modeling and characterization of plants have been undertaken by the research community with considerable success \citep{Okura2022}. Among the tasks that enable organ-based trait estimation are semantic segmentation and instance segmentation \citep{Schunck2021pheno4d, Godin2022}. In semantic segmentation, each point on the 3D surface is classified into one semantic category, such as leaf, branch, flower, or fruit. In instance segmentation, each point is assigned to a specific instance of an organ.

As a dimensionality reduction technique, t-distributed stochastic neighbor embedding (t-SNE) \citep{van2008visualizing} is widely used for visualization of high-dimensional data in a low-dimensional space of two or three dimensions. It is a nonlinear dimensionality reduction method which aims to keep similar data points close in the lower-dimentional space and to preserve the local structure of the data. Traditionally, t-SNE is used only for visualization purposes. However, in this paper, in addition to being a visualization tool for plant structures in various scales, t-SNE is demonstrated to be a suitable gateway for processing and clustering 3D point clouds in 2D space. 

t-SNE has many capabilities suited to inspecting and processing 3D plant models in 2D space. One rationalization of the application of t-SNE to 3D plant data is its manifold assumption, which holds that the high-dimensional data points are inherently organized to lie on lower-dimensional manifolds \citep{van2008visualizing}. This assumption resonates with the fact that, for certain plants species, much of the plant surface lies instrinsincly on one dimensional (thin branches, petioles) or two dimensional (leaf blades, petals) manifolds. t-SNE is known to be capable of capturing the local geometry of high-dimensional data, thus unwrapping structures that lie on different low-dimensional manifolds. Thus, t-SNE makes it possible to visualize the complete unoccluded structure of the 3D plant via a single 2D view (Fig. \ref{fig:tsneVisTobacco}). Furthermore, t-SNE also reveals both the local shapes of the parts and also the global organization of the data in the form of clusters at multiple scales in the 2D space \citep{van2008visualizing}.

As stated, t-SNE produces a 2D point set, with natural clusters and preserved local shape information, that can be processed for information extraction. In this paper, baseline methods that operate on the 2D points embedded by t-SNE are proposed for the applications of semantic and instance segmentation of 3D plants. The semantic segmentation is performed through extraction of superpoints in the 2D space. The initial partitioning is based on the natural clusters formed by t-SNE. The clusters are then further partitioned through 2D shape characteristics and 2D spectral clustering in order to ensure convexity of the 2D segments. The final superpoint labels are then mapped to 3D and each superpoint is described by simple geometric features for classification. Instance segmentation is applied only to the points classified as "Leaf". Leaf points are embedded to 2D by t-SNE and partitioned into individual leaves through unsupervised 2D Euclidean clustering.

The contributions of this work can be listed as follows:
\begin{itemize}
\item To the best of the author's knowledge, this is the first attempt of using t-SNE for visualization of 3D point clouds of plants in 2D. t-SNE enables inspection of the parts of the entire 3D plant in a single 2D view.
\item Similarly, to the best of the authors's knowledge, 2D points embedded by t-SNE have not been processed for analysis of 3D point clouds. To this end, baseline methods for semantic and instance segmentation through the use of t-SNE are developed.
\item A novel method is proposed to organize 3D point clouds of plants in terms of superpoints via 2D processing of the points embedded by t-SNE.
\end{itemize}

\section{Related Work}
\label{sec:relatedwork}

In this section, previous works targeting semantic and instance segmentation of 3D plant models is discussed. Special emphasis is given to three aspects of the related work: 1) Clustering approaches for segmentation of 3D plant models, 2) Local surface features tailored to distinguish organ classes, 3) Processing of 3D plant models in 2D space.

Clustering is a common approach to separate individual organs following semantic classification of 3D points on the plant surface. Individual leaves are either spatially detached from each other or exhibit strong geometric discontinuity at locations of contact, while structural units such as main stem, branches and petioles are connected. Thus, clustering is usually applied to regions containing points identified as leaves. For example,  \citet{elnashef2019tensor} clustered leaf points through DBSCAN to determine regions corresponding to individual leaves. To deliniate single leaves in areas containing multiple overlapping leaves, \citet{Liu2020automatic} applied a version of DBSCAN that is based on manifold distance. \citet{Sun2020} developed a density-based clustering method and applied it to 3D points identified as cotton bolls for the purpose of detecting and counting individual cotton bolls.

Clustering can also be applied to the entire plant surface for an initial segmentation to be refined with further information. \citet{golbach2016validation} developed a segmentation method based on the breath-first flood-fill algorithm where close points are added to organ clusters iteratively. The flooding of an organ cluster stops whenever a measure of spread exceeds a threshold. \citet{Xia2015} applied the mean shift algorithm to extract initial regions of individual leaves, and then employed active contour models to determine boundaries of occluded leaves. \citet{Liu2018} used Euclidean distance and spectral clustering (SC) algorithms for segmenting 3D point clouds into elementary shape units as candidate plant organs. \citet{Hetroy2016} created a graph by using neighborhood information of 3D points of tree seedlings. Then, spectral clustering is applied to the graph for an initial partitioning. The procedure is repeated on the clusters to further segment them to leaves, petioles, and internodes. \citet{Godin2022} constructed a neighbourhood graph to extract local geometric and spectral features for semantic segmentation. Then, they used a quotient graph to simultaneously refine the semantic labels and segment organ instances.

As in the works of \citet{Li2018leaf, Duan2016, Sun2020}, an initial oversegmentation of the points can be obtained in the form of superpoints of a 3D point cloud. \citet{Li2018leaf} obtained an over-segmentation on plant point clouds to aggregate the segments into individual leaves via region growing. \citet{Duan2016} partitioned point clouds using Octree organization into several groups, which corresponded to a slight over-segmentation of the cloud. The groups were then manually merged into individual organs. \citet{Sun2020} used a voxel cloud connectivity segmentation (VCCS) algorithm to obtain an over-segmentation in the form of supervoxels. Supervoxels corresponding to bolls and branches of cotton plants were classified through SVM operating on region shape descriptors. \citet{Wang2020} organized the point cloud of a forest environment as a superpoint graph which is then processed by an unsupervised network architecture for both single tree isolation and leaf-wood classification. In their TreePartNet framework, \citet{Liu2021} devised coupled networks that learned a neural decomposition of the input point cloud of a tree into a set of small-scale clusters, and merged oversegmented clusters through a pairwise affinity network.

Clustering based on spatial proximity, connectivity, or point density provides a partitioning of the point cloud data, however classification of the regions into semantic categories, such as leaf blades, stems, petioles, flowers, etc., requires descriptive information. Such descriptive information can be defined by quantifying the local geometric structure around the point to be classified. \citet{Dey2012} computed point-ness, curve-ness, and surface-ness features from the eigenvalues of local covariance matrix, and used SVM to classify the organs of grapevines. For leaf-stem classification, \citet{li2013analyzing} applied Markov Random Fields on the neighborhood graph of points. The data and smoothness terms were defined based on curvature values calculated using the eigenvalues from local PCA analysis. \citet{elnashef2019tensor} extracted descriptors based on the first and second tensor locally calculated around each point for semantic classification of leaves and stems. Using a deep learning classifier \citet{Ziamtsov2019} first selected Fast Point Feature Histograms (FPFH) as the best performing local feature among various alternatives. Then, they evaluated deep learning, support vector machines (SVM), K-nearest neighbors (KNN), and random forest on FPFH as classifiers.

Processing of 3D point clouds in the 2D domain is another practice for plant segmentation, mainly due to the ease of applying powerful convolutional neural networks on data defined on regular grids. For point clouds reconstructed from multiple 2D color and/or depth images, 2D plant scenes are already available. For example, \citet{Shi2019} exploited the 2D image data acquired for multi-view 3D reconstruction of seedlings. They applied convolutional neural networks (CNN) to 2D images for semantic and instance segmentation. The predictions on 2D images were then combined through a voting strategy to label corresponding 3D points. Through an embedding network,  \citet{Liu2020automatic} extracted features from depth images of poplar seedling leaves acquired from an RGB-D camera. Then, they applied Mask R-CNN to RGB-D data to determine leaf areas. \citet{Majeed2020} developed a CNN-based segmentation network to segment foreground RGB images of apple trees into trunk, branch and trellis-wire categories. The foreground images were extracted with the help of 3D point cloud data acquired with a Kinect V2 sensor.

For point clouds for which there is no associated natural 2D representations, for example for LiDAR data, 2D rendering is performed virtually. In this case, a number of issues should be addressed, such as the rendering method, number of images to be rendered, the resolution of the images, and the angles of the projection. In order to accomplish semantic labeling of 3D point clouds of grapevines, \citet{Japes2018} generated 2D images, together with their associated labels, from multiple viewpoints around mesh representations of the clouds. They used these images to train a U-Net-based neural newtork. During inference, the prediction scores from 2D images were back projected to 3D. \citet{Itakura2018} converted the 3D plant model to volumetric form and projected it onto the plane above the model. The resulting binary image was segmented with watershed algorithm. Then, the partitions were shrinked to determine seed regions. The expansion of seed regions is performed in the 3D space to extract final leaf instances. \citet{Jin2018deep} extracted horizontal slices from 3D point clouds of maize sites and the slices were projected into images as viewed from various angles. A Faster R-CNN was trained on these images to detect individual maize plants. For crown segmentation for individual rubber trees from LiDAR data, \citet{Wang2019individual} applied voxelization to the point cloud, and generated frontal and lateral projections of point clouds guided by the voxelization. These images were used to train a faster R-CNN that learned to locate individual tree trunks. 

For the sake of completeness, deep learning approaches that aim to segment 3D plant point clouds are also mentioned in this section. Such techniques necessitate large amount of annotated data for training \citep{Chaudhury20203d}. As 3D plant data sets are created \citep{Conn2017high, Conn2017statistical, Schunck2021pheno4d, Dutagaci2020rose}, and 3D annotation tools are developed \citep{miao2021label3dmaize, Saeed2022}, application of 3D deep learning architectures to plant data is becoming feasable. \citet{Turgut2022segmentation} compared six point-based deep learning architectures for semantic segmentation of 3D rosebush models. \citet{Schunck2021pheno4d} constructed the Pheno4D multi-temporal data set of tomato and maize plants. They reported semantic and instance segmentation results obtined by PointNet \citep{qi2017pointnet}, PointNet++ \citep{qi2017pointnet2}, and  LatticeNet \citep{Rosu2022latticenet}. \citet{Morel2020} proposed a network based on PointNet and PointNet++ for semantic segmentation of virtual trees. \citet{Chaudhury2021transferring} analyzed the performance of PointNet++, which was trained with virtual plants, on 3D models of real plants. \citet{Boogaard2021boosting} incorporated spectral information together with spatial input to boost the segmentation performance of PointNet++ on cucumber plants. \citet{Turgut2022rosesegnet} introduced RoseSegNet, where attention-based modules were used to encode local point interactions. \citet{Du2022pst} proposed the plant segmentation transformer (PST), which operated on voxelized data and employed attention blocks to encode contextual information. They applied PST for both semantic and instance segmentation of rapeseed plants. \citet{Ghahremani2021deep} developed Pattern-Net to partition the point clouds of spring wheat plants into ear and non-ear categories. \citet{Li2022plantnet} proposed PlantNet, which simultaneously achieved semantic and instance segmentation of plant models of various species, including tobacco, tomato, and sorghum \citep{Conn2017high, Conn2017statistical}. Targeting the same data set, \citet{Li2022psegnet} designed and trained PSegNet, which included three novel modules for feature extraction, feature fusion and attention. \citet{Li2022automatic} developed DeepSeg3DMaize to partition point clouds of maize plants into stem and leaf categories, and extract organ instances. By employing a class-dependent sampling strategy, \citet{Boogaard2022} trained PointNet++ to obtain a semantic segmentation with an enriched set of categories including “stem,” “petiole,” “leaf,” “growing point,” “node,” “ovary,” “tendril” and “non-plant”. \citet{Wang2022} proposed PartNet, which was designed to recursively partition 3D point cloud of lettuce plants into leaf instances in a top-down manner.

The current work proposes the use of t-SNE as a gateway for application of many of the methods mentioned above. The embedded 2D points and their natural clusters provided by t-SNE enable the implementation of 2D counterparts of those 3D techniques, such as superpoint extraction, split-and-merge approaches, graph-based methods, 2D deep learning architectures, etc. Also, the clusters extracted by t-SNE can be encoded and labeled by well-studied 2D or 3D shape descriptors, whether hand-crafted or extracted through deep neural networks. In this work, we demonstrate the application of superpoint extraction in the 2D embedded space and description of them through local geometric features for the purpose of semantic segmentation of 3D plant models.

\section{t-distributed stochastic neighbor embedding}
\label{sec:t-SNE}

t-distributed stochastic neighbor embedding (t-SNE) is a visualization technique that re-organizes high-dimensional data in 2D or 3D space \citep{van2008visualizing}. To achieve a global organization in the form of clusters, t-SNE explores the implicit structure of all the data via random walks on neighborhood graphs. 

In this paper, t-SNE algorithm as introduced in \citep{van2008visualizing} is utilized for embedding of 3D plant point clouds in 2D. Given a 3D point set $\mathcal{X}$ of a plant model, the specific objective of this study is to map the points in set $\mathcal{X}$ to a set $\mathcal{Y}$ of 2D points. To embed the points in $\mathcal{X}$ into 2D space, t-SNE attempts to minimize the Kullback-Leibler divergence between a Gaussian distribution of the points in the original space and a Student t-distribution of points in the target space. Kullback-Leibler divergence  $KL(P \mid\mid Q)$ between the joint probability distribution, $P$, in the high-dimensional space and the joint probability distribution, $Q$, in the low-dimensional space is given as

\begin{equation}
KL(P \mid\mid Q) =  \sum_{j}{\sum_{i \neq j}{p_{ij}\log{\frac{p_{ij}}{q_{ij}}}}}
\label{eq:KL}
\end{equation}

\noindent where $p_{ij}$ and $q_{ij}$ correspond to joint probabilites of point pair $(i,j)$ under the distributions $P$ and $Q$, respectively.

The distance $d(x_i,x_j)$ between each pair of points $x_i$ and $x_j$ in the set $\mathcal{X}$ is calculated as a measure of similarity between the points. In this work, Euclidean distance is selected as the similarity measure. This similarity is used to estimate the conditional probability $p_{j\mid i}$, which can be interpreted as the probability of $x_j$ occuring around $x_i$ under a Gaussian distribution centered at $x_i$, and with variance $\sigma_i$. The conditional probability $p_{j\mid i}$ is estimated as

\begin{equation}
p_{j\mid i} = \frac{exp( - \|x_i - x_j \|^2/2\sigma_i^2)}{\sum_{k \neq i}{exp( - \|x_i - x_k \|^2/2\sigma_i^2)}}
\end{equation}

with $p_{i\mid i}$ set to 0. As stated, the conditional probability distribution, denoted as $P_i$, is assumed to be Gaussian and specific to the datapoint $x_i$. $P_i$ embodies the distribution of all other data points in the set around $x_i$. t-SNE searches for the value of the varaiance $\sigma_i$ such that the Shannon entropy $H(P_i)$ of $P_i$ satisfies

\begin{equation}
Perp(P_i) = 2^{H(P_i)}
\label{eq:perp}
\end{equation}

The perplexity $Perp(P_i)$ is a user-defined parameter. The Shannon entropy of $P_i$ measured in bits is calculated as

\begin{equation}
H(P_i) = - \sum_{j}{p_{j\mid i} \log_2 (p_{j\mid i})}
\end{equation}

The joint probability $p_{ij}$ is calculated by symmetrizing the conditional probabilities:

\begin{equation}
p_{ij} = \frac{p_{j\mid i}+p_{i\mid j}}{2N}
\end{equation}

where $N$ is the number of points in set $\mathcal{X}$.

The probability model $q_{ij}$ of the distribution of the distances between points $y_i$ and $y_j$ in the low dimensional space is set to be a Student t-distribution with one degree of freedom. The joint probabilities $q_{ij}$ are defined as

\begin{equation}
q_{ij} = \frac{(1+\|y_i - y_j \|^2)^{-1}}{\sum_{k}{\sum_{l \neq k}{(1+\|y_k - y_l \|^2)^{-1}}}}
\end{equation}

with $q_{ii}$ set to 0. To minimize the Kullback-Leibler divergence between $P$ and $Q$ given in Eq. \ref{eq:KL}, t-SNE uses a gradient descent procedure. The details can be found in \cite{van2008visualizing}. The resulting Kullback-Leibler divergence can be regarded as the "loss" value measuring the divergence between pairwise similarities of points in the high-dimensional space and of the embedded points in the low-dimensional space. 

t-SNE computes the conditional probability distribution $P_i$ for each point $x_i$ by setting a constant $Perplexity$ value. Thus, it adaptively adjusts the effective neighborhood size depending on local point density around each point \citep{van2008visualizing}. For larger or denser point sets larger values of $Perplexity$ is preferred for better capture of the data in the low dimensional space. The dependency of the embedding to the $Perplexity$ requires user interaction to search for the appropriate value for interpreting high-dimensional data whose nature is relatively unknown and to be discovered. However, for 3D plant point clouds,  the point resolution, and the expected physical dimensions of structures of interest are known beforehand, allowing the $Perplexity$ parameter to be adjusted to reveal the desired level of organization.

\section{t-SNE for visualization of plant point clouds}
\label{sec:visualization}

In this section, visualization of 3D plant point clouds in 2D is suggested as an application of dimension reduction via t-SNE. Visualization in 3D requires rotation of the model in 3D for allowing the viewer to explore various structures from different views. t-SNE provides a one-shot visualization where all components are flattenned and laid out in 2D. Given a point cloud $\mathcal{X}$ composed of $N$ 3D points, the point set $\mathcal{Y}$ including corresponding $N$ 2D points are obtained through t-SNE as described in \cite{van2008visualizing}. The parameters involved in the operation are the $Perplexity$ and the number of points $N$, as determined by the size of the plant and the point density. Depending on both paremeters, the 2D embedding presents various layouts of the structures of the plant.

As stated before, t-SNE assumes that the high-dimensional data set is closer to a lower-dimensional manifold. It attempts to probe such manifold structures in the data and organize them in clusters.  Many plant species have geometric structures that inherently follow lower-dimensional manifolds. The most notable example is a broad leaf blade whose points spread on a 2D continuous and smooth manifold.  Thus, t-SNE has the potential to separate the plant data at sharp discontinuities in terms of orientation and group points together that lie on the same manifold. Furthermore, t-SNE embeds the points in each cluster onto 2D space such that close points stay together, resulting in a faithfully flattened view of the clustered structures.

\begin{figure}
	\centering
	\fbox{\includegraphics[scale=0.55]{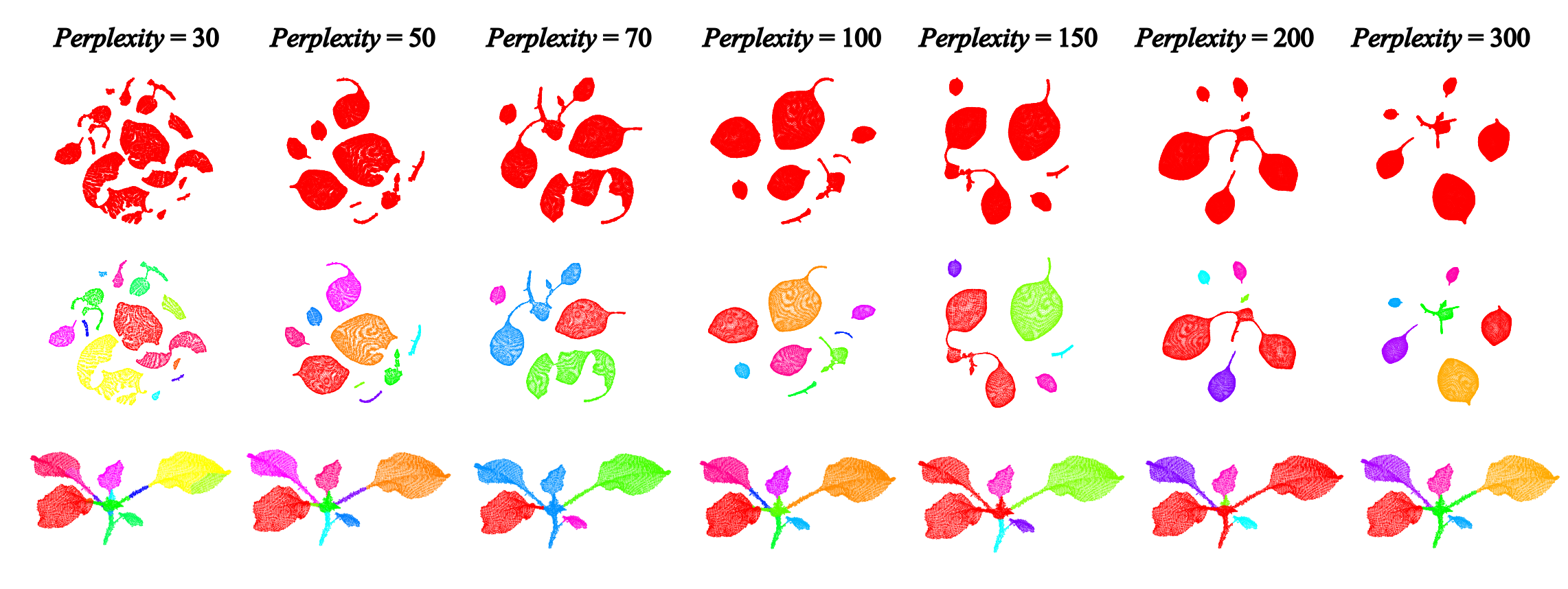}}
	\caption{First row: t-SNE embeddings of a tobacco plant model from the data set of \cite{Conn2017high} with varying $Perplexity$. Second row: t-SNE embeddings where the data is clustered through Euclidean clustering in 2D, and each cluster is rendered in a different color. Third row: The same 3D point cloud colored according to the clusters obtained from 2D embeddings. }
	\label{fig:tsneVisTobacco}
\end{figure}

\begin{figure}
	\centering
	\fbox{\includegraphics[scale=0.6]{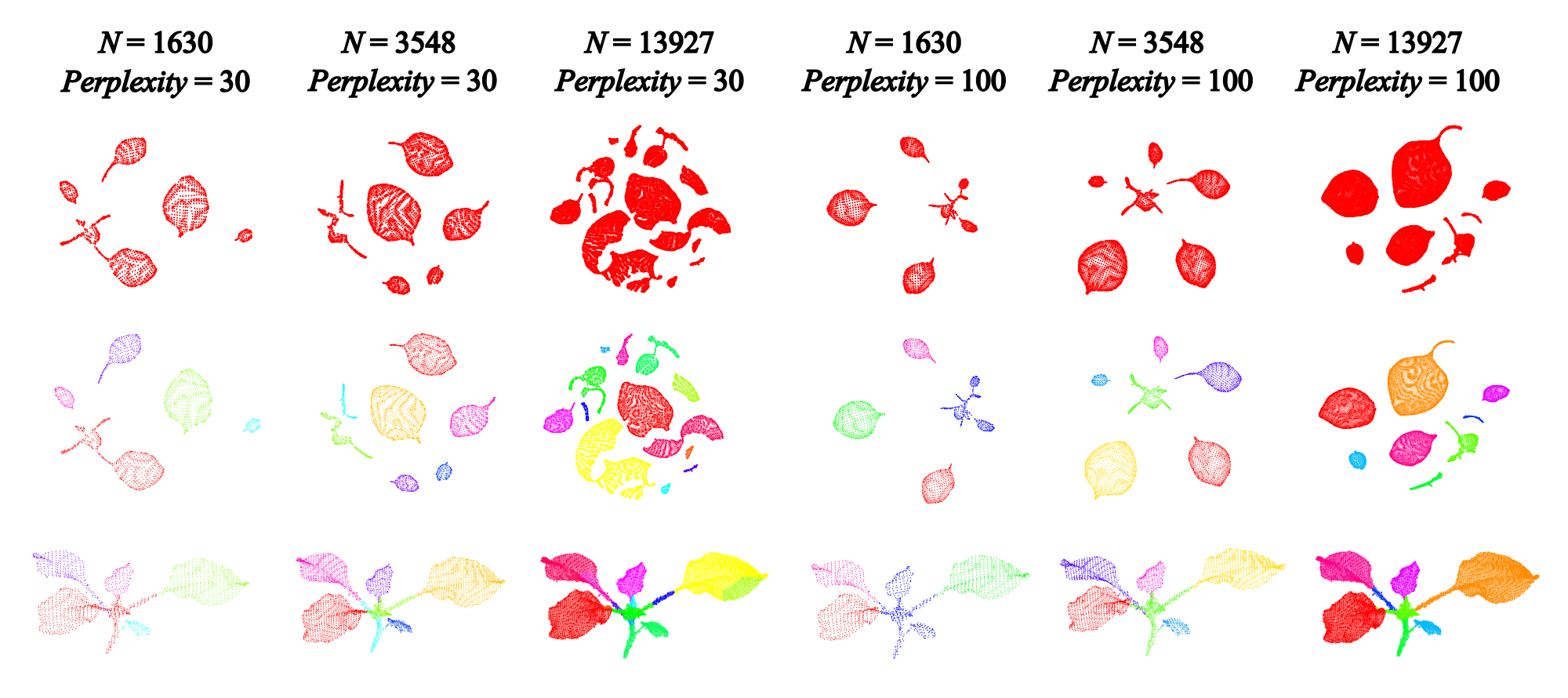}}
	\caption{First row: t-SNE embeddings of a tobacco plant model from the data set of \cite{Conn2017high} for three different resolutions and two $Perplexity$ values. Second row: t-SNE embeddings where the data is clustered through Euclidean clustering in 2D, and each cluster is rendered in a different color. Third row: The same 3D point cloud colored according to the clusters obtained from 2D embeddings.}
	\label{fig:tsneVisTobaccores}
\end{figure}

\begin{figure}
	\centering
	\fbox{\includegraphics[scale=0.62]{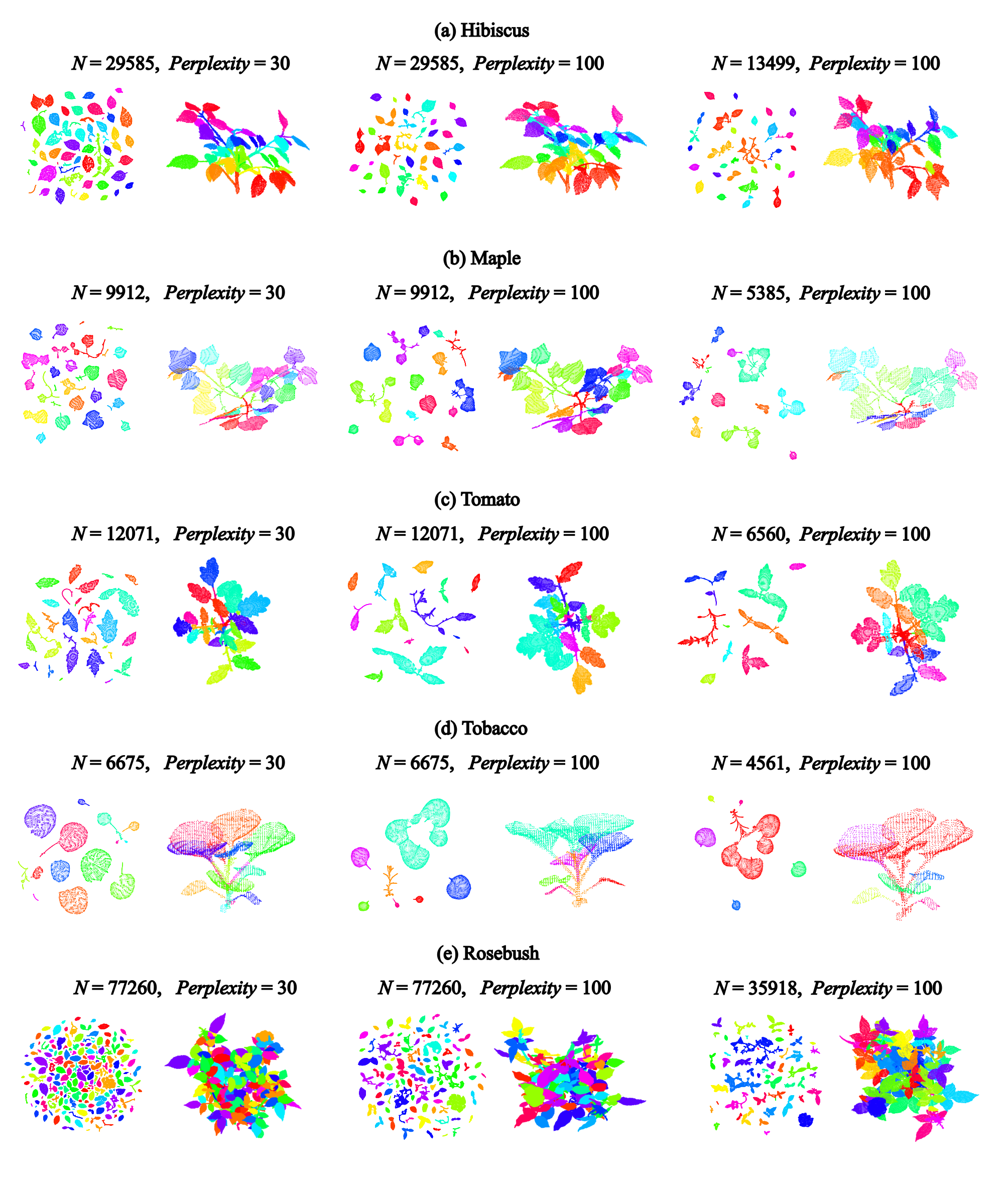}}
	\caption{t-SNE embeddings of various plant point clouds. 2D data is clustered through Euclidean clustering. Each cluster is rendered in a different color. 3D point clouds are colored according to the clusters obtained from 2D embeddings.}
	\label{fig:tsneVisPlants}
\end{figure}

A tobacco plant with few organs and a simple architecture is chosen for the purpose of illustration of 3D plant visualization through t-SNE (Figs. \ref{fig:tsneVisTobacco} \& \ref{fig:tsneVisTobaccores}). The 3D model belongs to the data set provided by \citet{Conn2017high}. Visualizations of more complicated plants are presented in Fig. \ref{fig:tsneVisPlants} The 3D point cloud of the tobacco plant is first subsampled to have $N=13927$ points. In the first row of Fig. \ref{fig:tsneVisTobacco}, 2D embeddings of the point cloud through t-SNE are given. The $Perplexity$ value is varied to observe its effect on the resulting layout of the embedded points in 2D. In the second row of Fig. \ref{fig:tsneVisTobacco}, the clusters of the 2D point sets are rendered with different colors. The clusters are obtained through Euclidean clustering, which is detailed in Section \ref{sec:superpoint}. The third row of Fig. \ref{fig:tsneVisTobacco} illustrates the same 3D point cloud colored according to the clusters obtained from 2D embeddings. Although the $Perplexity$ parameter plays a significant role in the forming of clusters, a large range of $Perplexity$ values (between 70 and 300 in the example shown in Fig. \ref{fig:tsneVisTobacco}), produces visually satisfactory embeddings that form natural clusters of the plant and reveal unrolled 2D shapes of the structures. Thus, once the desired level of organization is obtained, a fine tuning of the $Perplexity$ value is not required.

At this level of point density ($N=13927$), when $Perplexity$ is 30, t-SNE tends to segment individual leaves at orientation discontinuities. When $Perplexity$ is lower than 100, large leaves can be cut at orientation discontinuities. When the $Perplexity$ is 100 and above, each individual leaf is entirely encompassed by a single cluster. The boundaries of the embedded leaves are in accordance with the corresponding 3D boundaries of leaves. This is a significant strength of t-SNE since it allows to render the shape of each leaf close to the shape if the leaf was flattenned physically. Even small leaves that are barely visible in 3D and that require careful exploring to be noticed become apparent in 2D t-SNE embedding. From Fig. \ref{fig:tsneVisTobacco}, it can be observed that the small leaf at the branching point is observable as a small blob in the shape of a small leaf in all the 2D renderings. Through a single view, the observer can count the leaves (possibly also the petioles and internodes) and assess their shapes. 

Another observation is that, with increasing $Perplexity$ value, the 2D embedding reveals more about the plant architecture. Connected leaves start to stay connected in 2D through their corresponding petioles. Although not particularly significant for this example, this property of t-SNE allows grouping and rendering of substructures in 3D plant models with complicated architectures. Examples demonstrating this utility of t-SNE can be found in Fig. \ref{fig:tsneVisPlants}.

To observe the effect of point density, as quantified with number of points, $N$ in the point cloud, 2D embeddings of the same tobacco plant with three different subsampling rates are given in Fig. \ref{fig:tsneVisTobaccores}. The trade off between efficient shape summarization and the retaining of detail is common to all 3D point cloud processing methods. The main disadvantage of low point resolution is the loss of discerning data for small structures. For example, although still visibile, the small leaf at the branching point loses its discriminating shape in the embeddings with $N = 1630$ and $N = 3548$.  However, when the number of points covering individual leaves drop, the tendency of t-SNE to partition individual leaves is no longer present at low $Perplexity$ values. With low resolution, the leaf shape and the natural clusters are still observed. The branch structure appears as more connected in the 2D embeddings as compared to those from high resolution. As the resolution increases, individual branches and petioles appear as separate clusters.

\section{t-SNE for superpoint extraction from 3D point clouds}
\label{sec:superpoint}

As discussed in \ref{sec:relatedwork}, segmentation approaches following a split-and-merge spirit are quite common for processing 3D point clouds of plants. Oversegmentation in the form of superpoints, which correspond to groups of points with similar local geometry, simplifies subsequent processing via treating the superpoint as a single and homogeneous unit \citep{landrieu2018large}. In this section, a novel procedure that exploits t-SNE and generates a partitioning of the point cloud in the form of superpoints is proposed. The effectiveness of the proposed method is a demonstration of potential applications of t-SNE for 2D processing of 3D plant point clouds. The procedure starts with the clusters formed by t-SNE, then partitions each cluster further according to a number of criteria. Once the 2D superpoints are extracted, the superpoint labels are mapped to the 3D cloud.

\begin{figure}
	\centering
	\fbox{\includegraphics[scale=0.55]{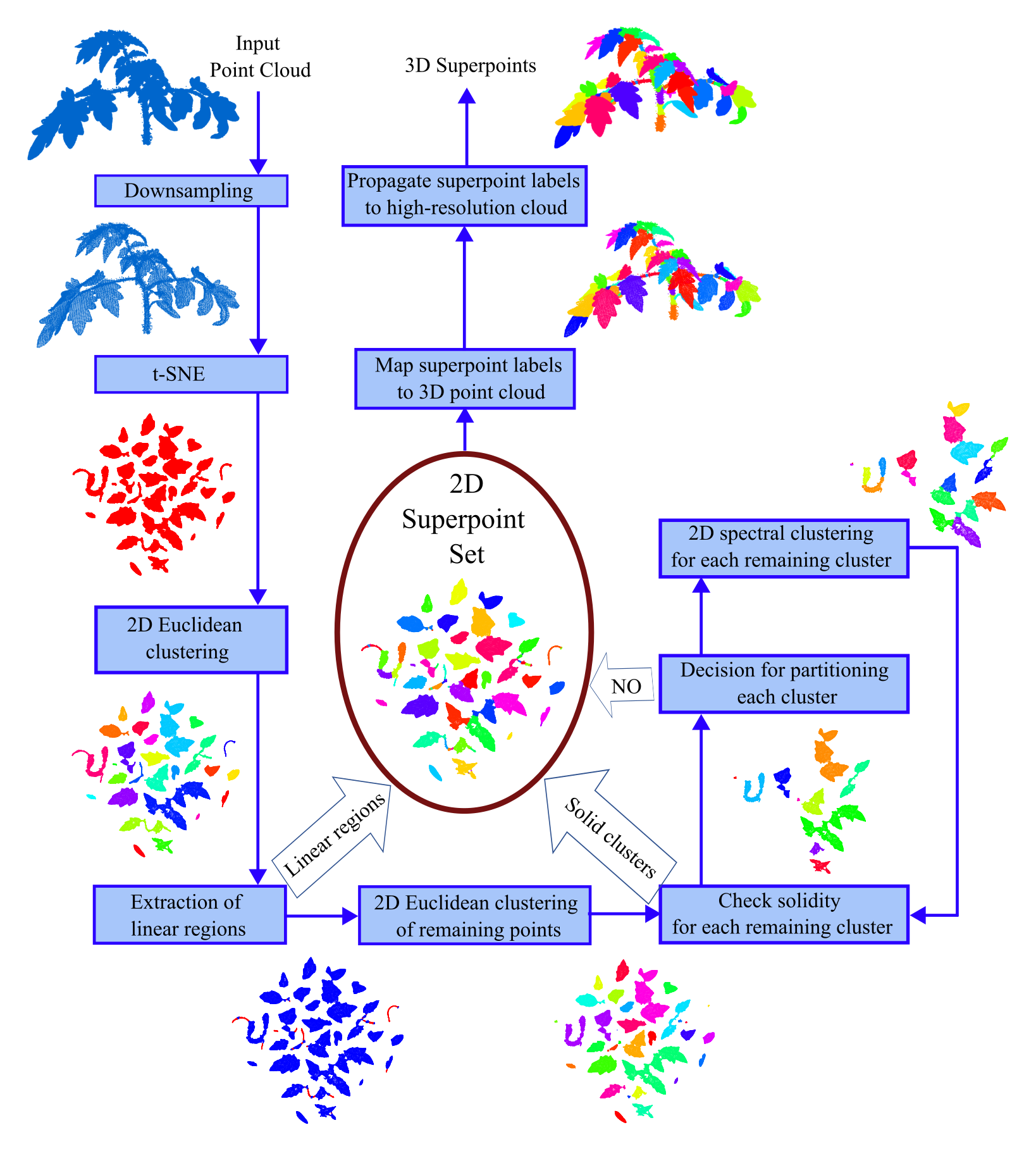}}
	\caption{Procedure to extract superpoints from 2D points embedded through t-SNE.}
	\label{fig:superFlowChart}
\end{figure}

The procedure of extraction of superpoints from the 2D embedding yielded by t-SNE is given in Fig. \ref{fig:superFlowChart}. The input point cloud is downsampled via voxel grid average filtering before t-SNE is applied. The size of the voxel cube in the grid determines point density and is denoted as $v$. Throughout this study, $v$ is set to be 1 to ensure adequate point density to represent small leaves and, at the same time, low computational demand. If the number of points in the downsampled set is less than 1000, then $v$ is decreased so that the number of points is close to 1000. This condition allows t-SNE to appropriately process small seedlings. Once the downsampled set of 3D points are mapped to 2D via t-SNE, the rest of the algorithm operates entirely in the 2D space until the superpoint labels are determined. The steps of superpoints extraction in 2D are summarized in Fig. \ref{fig:superFlowChart} with demonstrative examples, and are detailed as follows:

\begin{enumerate}

\item \textbf{2D Euclidean clustering:} First step in 2D is to apply Euclidean clustering to the embedded points generated by t-SNE. Euclidean clustering produces clusters such that, between any two points in a cluster, there is a path that remains in the cluster, and the distance between any subsequent points in the path is smaller than a threshold \citep{RusuDoctoralDissertation}. This threshold is denoted as $d_E$, and is one of the adjustable parameters of the superpoint extraction pipeline.

\item \textbf{Extraction of linear regions:} For each cluster obtained from the previous step, a local Principal Component Analysis (PCA) is performed to extract thin, line-like regions. The objective here is to detect structures that might correspond to thin branches and petioles and separate them from other structures to which they are connected in the 2D embedding. For each point in each cluster, a local neighborhood of radius $r_E$ is defined, and PCA is applied to the neighboring points to extract two principal directions and corresponding eigenvalues. $r_E$ is always set to be equal to $2d_E$ throughout the experiments in this study. The ratio of the smaller eigenvalue to the larger is an indicator of the linearity at the locality. A small threshold $T_E = 0.1$ is chosen throughout all the experiments in order to extract only strictly line-like points. The connencted components of the points that have eigenvalue ratio smaller than $T_E$ are formed and these regions are included in the 2D Superpoint Set.

\item  \textbf{2D Euclidean clustering of remaining points:} After the removal of the line-like regions, 2D Euclidean clustering is applied to the remaining points, and the clusters are updated. The structures connected by branches or petioles in the 2D embedding are thus disconnected through this operation. 

\item  \textbf{Check solidity for each remaining cluster:} For each cluster in the updated cluster set, which are not labeled as superpoints yet, the solidity of the cluster is calculated. Solidity is a measure of the convexity of a shape. In 2D, it is described as the ratio of the area of the shape to its convex area. The boundary points of each cluster are extracted and converted to a polygon. Through this polygon representation, the areas of the shape and its convex hull are calculated. The clusters with a solidity larger than $s_E = 0.8$ are deemed to be convex and added to the Superpoint Set. The remaining clusters are further examined for partitionings, since their concavity is an indicator that the cluster might include merged structures, such as merged individual leaves.

\item  \textbf{Decision for partitioning each cluster:} For the remaining clusters whose solidity values are lower than $s_E$, spectral analysis is performed to decide whether or not to partition the cluster further. Spectral clustering is a graph-based clustering method, which splits the graph in $K_s$ partitions using the $K_s$ eigenvectors of the Laplacian matrix of the graph \citep{Shi2000normalized, Ng2001spectral}. The number of clusters $K_s$ is usually assumed to be equal to the number of eigenvalues of the Laplacian matrix that are close to 0. The algorithm as described in \cite{Ng2001spectral} is applied to the 2D points in a cluster, and $K_s$ is estimated to be equal to the number of eigenvalues that are close to zero. The criterion to be close to zero is chosen to be being smaller than $T_s = 0.0005$ throughout the experiments. If $K_s$ is equal to 1, indicating that the points in the cluster forms a single cluster, it is decided that no further partitioning is needed, and the cluster is added to the Superpoint Set.

\item  \textbf{2D spectral clustering for each remaining cluster:} The clusters for which $K_s$ is greater than 1 are partitioned into $K_s$ sub-clusters through spectral clustering. The cluster is replaced by the sub-clusters. The set of remaining clusters, which are not yet assigned to superpoint labels, are thus updated.

\item  \textbf{Iterate through Steps 4 to 7:} The updated set of clusters are reprocessed through Steps 4 to 7 as long as the set of unlabeled clusters is non-empty.

\end{enumerate}

An example to the resulting set of 2D superpoints is given in Fig. \ref{fig:superFlowChart}, where each superpoint is rendered with a different color. The superpoint labels are mapped to the corresponding points in 3D, then propagated to the high-resolution point cloud through nearest neighbor interpolation. Most of the superpoints are entirely included in a single organ, representing a single leaf, part of a single leaf, or part of a branch. As a consequence, subsequent segmentation algorithms can operate on the superpoints treating each as an undivided, homogeneous entity.

\section{Semantic segmentation}
\label{sec:semantic}

In this section, the utilization of superpoints extracted in the 2D space through t-SNE is demonstrated for semantic segmentation. Each superpoint is treated as a unity and assigned a single feature vector describing the superpoint. Notice that, a wide variety of descriptors can be used for representing the 2D or 3D point set of each superpoint, including deep features. In this study, geometric properties of spherical neighborhoods of 3D points were selected as baseline descriptors \citep{thomas2018semantic}. 

Let the original point cloud be described as a set of 3D points $\mathcal{X}$, where a point $x_i \in \mathbb{R}^3$ is represented with its 3D coordinates. A neighborhood of radius $r$ is defined around each point $x_i$ as $\mathcal{N}_i^{r} = \{x \in \mathcal{X} | \| x - x_i \| < r  \} $, where $\| \cdot \| $ is the Euclidean norm. Let the eigenvalues of the covariance matrix $C_i^r$ of the 3D points in $\mathcal{N}_i^{r}$ be denoted as $\lambda_1 > \lambda_2 > \lambda_3$. Using these eigenvalues, the following list of shape descriptors of the local neighborhood  $\mathcal{N}_i^{r}$ can be defined \citep{thomas2018semantic}:

\begin{itemize}
\item Anisotropy:   $f_{i,1}^{r} = \frac{\lambda_1 - \lambda_3}{\lambda_1}$
\item Planarity:   $f_{i,2}^{r} = \frac{\lambda_2 - \lambda_3}{\lambda_1}$
\item Sphericity:   $f_{i,3}^{r} = \frac{\lambda_3}{\lambda_1}$
\end{itemize}

In order to take into account various sizes of neighborhoods, features are extracted setting the radius to 3 different values, $r = 4, 5, 6$. The radius values are selected such that they cover the thickness of various points at branches. For each point in $x_i$ the set $\mathcal{X}$ a feature vector $F_i$ of dimension 9 is calculated. 

Let the set of superpoints be denoted as $\mathcal{SP}$, where superpoint $\mathcal{S}_n \in \mathcal{SP}$ is a subset of $\mathcal{X}$. All the 3D points in a single superpoint are suppossed to belong to the same semantic category. To describe the superpoint  $\mathcal{S}_n$, the mean of the feature vectors of the points in $\mathcal{S}_n$ is calculated:

\begin{equation}
F_n = \frac{1}{N_n} \sum_{x_i \in \mathcal{S}_n}{F_i}
\end{equation}

where $N_n$ is the number of 3D points in superpoint $\mathcal{S}_n$. The descriptors of superpoints are then used to classify each superpoint into one of two categories: "Leaf" and "Stem". A linear Support Vector Machine (SVM) classifier is trained on a training set of plant models. For the training plants, the label of each superpoint is assigned as the majority ground truth category of the 3D points in the superpoint. At inference stage, after the label of a superpoint is predicted by SVM all the 3D points in the superpoint is assigned to the same label. 

\section{Instance segmentation}
\label{sec:instance}

Instance segmentation can be performed on the entire unlabeled point cloud, or proceed semantic segmentation. In the latter case, instance segmentation is applied only to the points labeled with a single category. 2D embedding by t-SNE allows for both strategies to be applied. Shape information is preserved in the 2D space to delineate individual plant organs, e.g., employ various leaf instance segmentation techniques that operate in 2D \citep{scharr2016leaf}, including deep neural networks \citep{gu2022review}.

In this study, Euclidean clustering is applied to the embedded 2D points already classified as leaves by a semantic segmentation algorithm. The objective is to demonstrate that, given the appropriate parameters for point resolution and $Perplexity$, the clusters formed by the unsupervised t-SNE algorithm significantly overlaps with individual leaves. Even with this naive approach based solely on t-SNE clusters, the performance of instance segmentation with t-SNE clusters is shown to be on par with a deep learning approach (see Section \ref{sec:results}). The accuracy can be significantly improved by top-down (partitioning merged leaves) or bottom-up (merging torn leaves) without leaving the 2D domain.

Let a semantic segmentation algorithm label all 3D points in $\mathcal{X}$ with a set of labels including "Leaf". Let $\mathcal{L} \subset \mathcal{X}$ be the set of points labeled as "Leaf". First, $\mathcal{L}$ is downsampled using voxel grid average filtering with voxel size parameter $v$. In this work, $v$ is selected to be equal to 1, to ensure that small leaves are discernible. Then, the downsampled set $\mathcal{L}_s$ is reduced to 2D via t-SNE algorithm resulting in the 2D point set $\mathcal{Y}_s$. Euclidean clustering \citep{RusuDoctoralDissertation} is applied to $\mathcal{Y}_s$ to obtain clusters $L_k$ for $k= 1, 2,...,K$ and label each point with its corresponding cluster index $k$. Each cluster index is assummed to correspond to a single leaf. The labels of 2D points are mapped to 3D counterparts in $\mathcal{L}_s$, then propagated via nearest-neighbor interpolation to the high-resolution point set $\mathcal{L}$.

\section{Results}
\label{sec:results}

The proposed methods were applied to the tomato point clouds of the Pheno4D data set \citep{Schunck2021pheno4d}. The Pheno4D data set contains point clouds of seven tomato plants acquired by a laser scanning system at various time instances. 11 point clouds of each tomato plant were labeled, resulting in 77 labeled point clouds. The labels correspond to "Leaf", "Stem", and "Ground".  For this study, the points corresponding to the "Ground" were removed. In the Pheno4D data set, instance labels are also provided; i.e. each separate leaf is labeled with a distinct identity number.

\subsection{Semantic Segmentation}

For semantic segmentation, Intersection-over-Union (IoU) per class and mean Intersection-over-Union (mIoU) are used as performance measures. Following the experimental setup described in \cite{Schunck2021pheno4d}, 55 point clouds of the first five plants were used for training and the remaining two plants with 22 point clouds were reserved for testing. For the current study, the training set of plants was further partitioned into training and validation in a 4 to 1 ratio for determination of best-performing $Perplexity$ value and the threshold $d_E$ for Euclidean clustering. Table \ref{tab:parameters} gives the set of values of parameters with which 5-fold validation experiments were conducted for semantic segmentation. The parameter value that gives the best average mIoU is indicated in bold type. 

\begin{table}[!h]
	\caption{Parameters for 5-fold validation experiments for semantic segmentation. Best-performing parameters are indicated in bold type. 55 point clouds of five plants were used for training and validation sets partitioned in a 4 to 1 ratio.}
	\centering
	\begin{tabular}{l c}
		\toprule
		Parameter     & Set of values      \\
		\midrule
		$Perplexity$    & 20,  \textbf{30}, 40, 50   \\
		$d_E$    & 1,  \textbf{2}, 3  \\
		\bottomrule
	\end{tabular}
	\label{tab:parameters}
\end{table}

Table \ref{tab:semantic} gives the quantitative results of the proposed semantic segmentation procedure as compared to the results reported in \citep{Schunck2021pheno4d}. The results were obtained with the best-performing parameters given in Table \ref{tab:parameters}. 22 point clouds of two plants were used as test data. \citet{Schunck2021pheno4d} had applied PointNet \citep{qi2017pointnet}, PointNet++ \citep{qi2017pointnet2}, and LatticeNet \citep{Rosu2022latticenet} for semantic segmentation of the plant point clouds into three classes: "Leaf", "Stem", and "Ground". The current study exludes the classification of ground points. The mIoU values are given over three classes, as reported in \cite{Schunck2021pheno4d}, together with mIoU values over "Leaf" and "Stem" classes only.

\begin{figure}[!th]
	\centering
	\fbox{\includegraphics[scale=0.6]{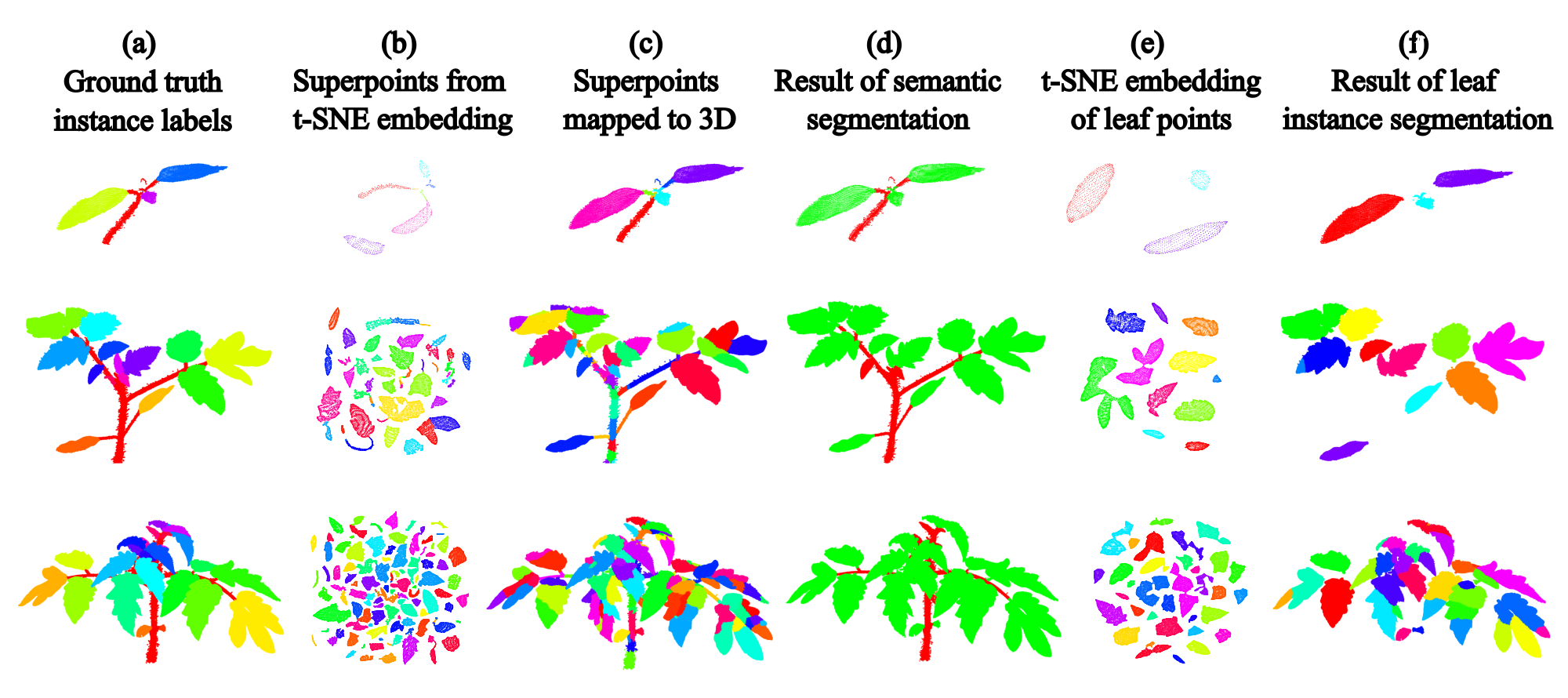}}
	\caption{Visual examples for semantic and instance segmentation on three point clouds corresponding to different time instances of growth of a tomato plant. (a) Ground truth instance labels, (b) Superpoints extracted from the 2D point set embedded by t-SNE, (c) Superpoints mapped to 3D and propagated to high resolution cloud, (d) Result of semantic segmentation, (e) Clusters in the t-SNE embedding applied to the points classified as "Leaf", (f) Result of leaf instance segmentation.}
	\label{fig:segResults}
\end{figure}

\begin{table}
	\caption{Semantic segmentation results given in IoU per class and mIoU on test tomato point clouds of Pheno4D data set \citep{Schunck2021pheno4d}. 22 point clouds of two plants were used as test data.}
	\centering
	\begin{tabular}{l c c c c c }
		\toprule
		  & IoU  &  IoU  &   IoU   & mIoU   & mIoU  \\
		& Leaf & Stem  &   Ground  & over 3 classes  & over 2 classes  \\
		\midrule
		PointNet \citep{qi2017pointnet} & 83.9  &  11.9 &  98.4   & 64.7  & 47.9  \\
		PointNet++ \citep{qi2017pointnet2}     &  88.8 & 19.6 & 98.6  & 68.4 &  54.2 \\
		LatticeNet \citep{Rosu2022latticenet}     & 98.7  & 86.9  & 99.8 &  95.1 &   92.8    \\
		\midrule
		Proposed    & 96.5   &  84.9 & N/A & N/A  & 90.7   \\
		\bottomrule
	\end{tabular}
	\label{tab:semantic}
\end{table}

As can be observed from Table \ref{tab:semantic}, the proposed method gives higher performance than PointNet and PointNet++, and is comparable to LatticeNet. Notice that, the superpoint extraction is performed in an unsupervised manner, and the descriptors for the superpoints are hand-crafted and simple. The results demonstrate that, processing 2D points embedded by t-SNE has the potential of keeping up with state-of-the-art methods operating in 3D space for semantic segmentation of plants.

Fig. \ref{fig:segResults} gives visual examples for semantic segmentation on three point clouds acquired at different stages of growth of a tomato plant \citep{Schunck2021pheno4d}. These clouds correspond to a small seedling, a medium-size tomato plant, and a larger plant with larger leaves. The respective size ratios are not kept in Fig. \ref{fig:segResults} for clear illustration of the smaller plants. The first column (Fig. \ref{fig:segResults}-(a)) shows the 3D point clouds with ground-truth instance labels of the leaves. In the second column (Fig. \ref{fig:segResults}-(b)) the t-SNE embedding with $Perplexity = 30$ and the extracted superpoints are given. For the seedling, small individual leaves remained in a single cluster, while for larger plants, large leaves were torn. However, the superpoint extraction algorithm managed to form large superpoints that include points from a single semantic class, and also discern small structures such as small leaves and petioles (Fig. \ref{fig:segResults}-(c)). The results of the semantic segmentation in Fig. \ref{fig:segResults}-(d) seems visually satisfactory apart from some small leaves classified as stem. Notice that these small leaves were successfully separated as superpoints, however the SVM classifier working with local geometric features labeled them as stem. There is room for improvement, as future work, through testing other shape descriptors and classifiers, or deep learning schemes for representing and classifying superpoints more effectively.

\subsection{Instance Segmentation}

For evaluation of instance segmentation performance, Symmetric Best Dice (SBD) \citep{scharr2016leaf} is used. In order to determine the best $Perplexity$ value for the particular data set, instance segmentation as described in Section \ref{sec:instance} is applied to the 55 point clouds of the first five tomato plants. The threshold for Euclidean clustering is set to $d_E = 2$. Points with ground-truth labeled as "Stem" are removed from the point clouds, and instance segmentation is applied only to the "Leaf" points. For determination of the $Perplexity$ value for instance segmentation, perfect prior semantic segmentation is assummed to decouple errors from the semantic segmentation algorithm. Table \ref{tab:perpInstance} gives the SBD performances with respect to the $Perplexity$ value. The highest SBD is yielded with $Perplexity$ value equal to 60.

\begin{table}[!h]
	\caption{Instance segmentation results given in SBD on 55 point clouds corresponding to the first five tomato plants in the Pheno4D data set \citep{Schunck2021pheno4d}. The performances with respect to the $Perplexity$ value are given. Perfect prior semantic segmentation is assummed.}
	\centering
	\begin{tabular}{l c c c c c c c }
		\toprule
		 $Perplexity$ & 20  & 30  &   40  & 50   & 60  & 70 & 80 \\
		\midrule
		SBD &  72.7 & 76.5 &  80.6    & 82.6  & 84.6  & 83.4 & 83.6 \\
		\bottomrule
	\end{tabular}
	\label{tab:perpInstance}
\end{table}

\begin{table}[!h]
	\caption{Instance segmentation results given in SBD on test tomato point clouds of Pheno4D data set \citep{Schunck2021pheno4d}. 22 point clouds of two plants were used as test data. }
	\centering
	\begin{tabular}{l c  }
		\toprule
		  & SBD \\
		\midrule
		PointNet \citep{qi2017pointnet} & 47.3  \\
		PointNet++ \citep{qi2017pointnet2}     &  56.1 \\
		LatticeNet \citep{Rosu2022latticenet}     & 74.2   \\
		\midrule
		Proposed    &  73.5 \\
		\bottomrule
	\end{tabular}
	\label{tab:instance}
\end{table}

In order to have a fair comparison with the instance segmentation results given in \cite{Schunck2021pheno4d}, the proposed instance segmentation method is tested on the 22 point clouds of the remaining two tomato plants. In this case, the proposed semantic segmentation is used as described in \ref{sec:semantic}, and then, instance segmentation through t-SNE is applied to the points classified as "Leaf". $Perplexity$ is set to be 60, as indicated by the results in Table \ref{tab:perpInstance}. The SBD value of the proposed method is given in Table \ref{tab:instance}, together with the performance values as reported in \cite{Schunck2021pheno4d}. With a simple unsupervised Euclidean clustering in the 2D space, and without further refinement, operating via t-SNE gives comparable instance segmentation performance (73.5\%) to that of LatticeNet (74.2\%). 

Fig. \ref{fig:segResults}-(e)\&(f) give t-SNE embeddings of points classified as leaf, and the clusters provided by Euclidean clustering mapped to 3D. For small and medium-size plants, the clusters mostly coincided entirely with a single leaf. For larger leaves, as shown in the third row, individual leaves with low convexity or with orientation discontinuities were torn into separate clusters. As a future work, strategies for merging these clusters through model-based techniques can be considered.

\section{Conclusion}

Through this work, t-SNE is discovered to be a powerful tool for visualization and processing of 3D point clouds of plants. It allows a single shot 2D view of the structures of the plant while preserving local shape information. Apart from visualization, t-SNE clusters plant data into natural partitions, leading to an initialization for successful segmentation methods. As a demonstration of this property, a novel superpoint extraction method that exclusively operates in 2D space is proposed. The effectiveness of this method is evaluated through semantic segmentation, where 90.7\% mIoU is obtained on the Pheno4D tomato data set. The performance of t-SNE ( 73.5\% SBD) via a simple unsupervised Euclidean clustering for instance segmentation is also shown to be on par with state-of-the-art 3D deep learning techniques.

\bibliographystyle{unsrtnat}
\bibliography{tSNEbib}

\end{document}